\documentclass[conference]{IEEEtran}
\IEEEoverridecommandlockouts
\usepackage{cite}
\usepackage{amsmath,amssymb,amsfonts}
\usepackage{algorithmic}
\usepackage{graphicx}
\usepackage{textcomp}
\usepackage{xcolor}
\usepackage{array}
\usepackage{booktabs}
\usepackage{balance}

\def\BibTeX{{\rm B\kern-.05em{\sc i\kern-.025em b}\kern-.08em
    T\kern-.1667em\lower.7ex\hbox{E}\kern-.125emX}}
\begin{document}

\title{Detection of Illicit Content on Online Marketplaces using Large Language Models}

\author{\IEEEauthorblockN{1\textsuperscript{st} Quoc Khoa Tran}
\IEEEauthorblockA{\textit{AiLECS Lab} \\
\textit{Monash University}\\
Melbourne, Australia \\
qtra0027@student.monash.edu}
\and
\IEEEauthorblockN{2\textsuperscript{nd} Thanh Thi Nguyen}
\IEEEauthorblockA{\textit{AiLECS Lab} \\
\textit{Monash University}\\
Melbourne, Australia \\
thanh.nguyen9@monash.edu\textsuperscript{*}\thanks{\textsuperscript{*}Corresponding author.}}
\and
\IEEEauthorblockN{3\textsuperscript{rd} Campbell Wilson}
\IEEEauthorblockA{\textit{AiLECS Lab} \\
\textit{Monash University}\\
Melbourne, Australia \\
campbell.wilson@monash.edu}
}

\maketitle

\begin{abstract}
Online marketplaces, while revolutionizing global commerce, have inadvertently facilitated the proliferation of illicit activities, including drug trafficking, counterfeit sales, and cybercrimes. Traditional content moderation methods such as manual reviews and rule-based automated systems struggle with scalability, dynamic obfuscation techniques, and multilingual content. Conventional machine learning models, though effective in simpler contexts, often falter when confronting the semantic complexities and linguistic nuances characteristic of illicit marketplace communications. This research investigates the efficacy of Large Language Models (LLMs), specifically Meta's Llama 3.2 and Google's Gemma 3, in detecting and classifying illicit online marketplace content using the multilingual DUTA10K dataset. Employing fine-tuning techniques such as Parameter-Efficient Fine-Tuning (PEFT) and quantization, these models were systematically benchmarked against a foundational transformer-based model (BERT) and traditional machine learning baselines (Support Vector Machines and Naive Bayes). Experimental results reveal a task-dependent advantage for LLMs. In binary classification (illicit vs. non-illicit), Llama 3.2 demonstrated performance comparable to traditional methods. However, for complex, imbalanced multi-class classification involving 40 specific illicit categories, Llama 3.2 significantly surpassed all baseline models. These findings offer substantial practical implications for enhancing online safety, equipping law enforcement agencies, e-commerce platforms, and cybersecurity specialists with more effective, scalable, and adaptive tools for illicit content detection and moderation.
\end{abstract}


\section{Introduction}
Online marketplaces have changed the way business is done around the world, but they have also become fertile ground for illegal activity \cite{RN53}. This makes it hard to keep things safe, regulated, and trustworthy \cite{RN55}. People use these sites to buy and sell illegal items and services such drugs, fake products, stolen data, firearms, and illegal services like hacking~\cite{RN114}. ``Platformization'' is when illegal actors copy real e-commerce sites. This makes it easier for them to get in and harder to find \cite{RN114}, which shows that we need to take a broad approach to stopping them.

The rise of these kinds of activities has serious effects on society, such as losing money, putting public health at risk, and making people less trusting of the digital economy. The fact that online platforms are anonymous and can be used all over the world makes these problems worse, making it harder for law enforcement to do their jobs and requiring detection systems that are effective, scalable, and adaptable \cite{RN55}.

But there are a lot of problems with traditional content control. It is not possible to scale up manual review, and rule-based systems have trouble with the changing ways that bad actors try to get around rules \cite{RN55}. This causes false negatives and positives, which are quite expensive. Conventional Machine Learning (ML) models often need a lot of feature engineering and do not work well with semantic complexity and multilingual information. This is a big problem because these marketplaces are worldwide. This ``arms race'' that keeps going needs smarter technologies.

Recent progress in AI has led to the creation of Large Language Models (LLMs), which have changed the way we perceive natural language. LLMs are highly effective at making predictions based on complicated context and language patterns that traditional approaches cannot understand. They are potential for finding a wide range of illegal content across languages since they can read and understand complicated text, including coded language, and adapt through fine-tuning \cite{guo2024moderating}. This goes from manual feature engineering to automated representation learning, and fine-tuning is a way to get specialized AI tools for safety.

Despite this promise, a critical gap exists in the rigorous comparative evaluation of the latest LLMs for complex, multilingual illicit content detection. There is a specific need to assess their performance on specialized, ``in-the-wild'' multilingual datasets like DUTA10K \cite{RN65, RN51}, as standard benchmarks often miss the unique linguistic traits of illicit online trade~\cite{RN55}. Furthermore, a comprehensive comparison of fine-tuned Llama~3.2 and Gemma 3 against traditional ML baselines and foundational transformers (like BERT) for both binary and granular multi-class classification in this specific context is underexplored. Such analysis is vital to quantify LLM advancements and assess fine-tuning efficacy.

This research primarily aims to evaluate and compare the performance of fine-tuned Llama 3.2 and Gemma 3 LLMs against selected traditional ML models and a BERT baseline for detecting and classifying illicit content using the DUTA10K multilingual dataset, across both multi-class and binary classification tasks. The key contributions include a systematic empirical evaluation of these LLMs on this challenging dataset; a direct comparative performance analysis against baselines for both classification task types; an investigation into the efficacy of Parameter-Efficient Fine-Tuning (PEFT) and quantization for this domain; and provision of insights into task-dependent model strengths to inform the development of more effective online safety tools.

The remainder of this paper is organized as follows: Section \ref{sec:related_work} reviews related work in illicit content detection and the application of ML models. Section \ref{sec:methodology} details the methodology, including the DUTA10K dataset, ethical considerations, the architectures of the LLMs and traditional models used, and the experimental setup. Section \ref{sec:results_discussions} presents and discusses the experimental results. Finally, Section \ref{sec:conclusions} concludes the paper, summarizes key findings, and outlines potential directions.

\section{Related Work}
\label{sec:related_work}

\subsection{Traditional Supervised Learning Approaches}
\label{sec:subsection_traditional}

Early research predominantly utilized traditional supervised learning algorithms. Support Vector Machine (SVM), logistic regression, decision trees, and naive Bayes classifiers were common choices. The study in \cite{RN30} introduced the Darknet Usage Text Addresses (DUTA) dataset, applying TF-IDF representation coupled with Logistic Regression, achieving a notable accuracy of 96.6\% and a macro F1-score of 93.7\% for categorizing darknet web textual content. This demonstrated the potential of classical ML with robust feature engineering. On the other hand, \cite{RN34} analyzed online consumer fraud using logistic regression and decision trees, reporting high precision (often \(>\)90\%) in distinguishing fraudulent from legitimate transactions. Similarly, \cite{sharif2020detecting} employed comparable methods, attaining strong performance in detecting suspicious text communications linked to illegal activities. While effective on specific tasks and often computationally efficient, these models heavily rely on manual feature engineering, which can be domain-specific and may not capture deeper semantic nuances or adapt well to new ways of hiding information.

\subsection{Unsupervised Learning Approaches}
\label{sec:subsection_unsupervised}
Given the frequent scarcity of labeled data for emerging illicit activities, unsupervised methods like clustering and topic modeling have been explored. The work in \cite{RN29} employed clustering algorithms to categorize drug-related activities on darknet platforms, successfully uncovering meaningful patterns despite inherent labeling challenges. Likewise, \cite{RN36} introduced an unsupervised ML pipeline combining clustering and Biterm Topic Modeling to detect and categorize suspicious darknet activities. While these approaches can highlight relevant clusters and topics without labeled data, they typically require manual validation and interpretation to ensure accuracy and assign meaning to discovered patterns, which can limit their scalability and applicability for automated moderation.

\subsection{Deep Learning Techniques}
\label{sec:subsection_deep}

Advancements in deep learning significantly enhanced illicit content detection capabilities. Convolutional Neural Networks (CNNs), effective for capturing local patterns, and Recurrent Neural Networks (RNNs), suited for sequential data like text, proved particularly useful. The study in \cite{RN16} utilized CNNs for detecting illegal wildlife trade through image analysis, achieving high generalization with F-scores between 0.75 and 0.95. In contrast, \cite{RN48} applied Long Short-Term Memory (LSTM) networks, a type of RNN, to identify and classify cybercrime-related content in sequential text data, demonstrating accuracy rates consistently above 90\%. In another work, \cite{RN11} highlighted CNN-based approaches for social media content classification, achieving remarkable precision and recall. These methods reduced the need for manual feature engineering by learning representations from data but could still be challenged by long-range dependencies and complex linguistic structures.

\subsection{Transformer-Based Models}
\label{sec:subsection_transformer}

The emergence of transformer-based models, notably BERT~\cite{RN67} and RoBERTa \cite{RN51}, marked a significant breakthrough in handling linguistic nuances and multilingual data. Their attention mechanisms allow for a more profound understanding of context. The study in \cite{RN20} demonstrated the efficacy of transformer-based methods for classifying drug trafficking activities on social media, achieving impressive accuracy exceeding 92\% by effectively interpreting slang and context. The work in \cite{RN19} extended that, reporting substantial improvements in capturing nuanced drug-related communications with high recall. Likewise, \cite{RN17} further utilized BERT for detecting harmful and illicit content online, noting improved context-awareness and semantic understanding compared to earlier methods. These models set a new benchmark but fine-tuning them can still be resource-intensive.

\subsection{Large Language Models (LLMs)}
\label{sec:subsection_LLMs}

More recent studies leveraging LLMs such as GPT variants have opened new avenues. A knowledge-prompted ChatGPT model was presented in \cite{RN15} to detect drug-related content on Instagram, significantly surpassing the performance of earlier models, with accuracies consistently above 95\%. Similarly, \cite{RN28} successfully applied GPT-based models to categorize cybercrime discussions within the Agora darknet marketplace dataset, achieving approximately 96\% accuracy despite challenges from linguistic obfuscation and dataset imbalance. The introduction of these even larger models highlights their potential for dynamic adaptation to evolving deceptive language. However, access to the most powerful proprietary LLMs can be restricted or costly, and their application to specific, multilingual datasets like DUTA10K with newer open models like Llama 3.2 and Gemma 3 warrants dedicated investigation.

\subsection{Research Gaps}
\label{sec:subsection_researchgap}

Despite these advancements, several research gaps persist, which this study aims to address. Firstly, there is a lack of comprehensive, direct comparative analyses between the latest generation of fine-tuned open LLMs (like Llama 3.2 and Gemma 3), traditional ML, earlier deep learning models, and foundational transformers like BERT, specifically for \textit{both} binary and fine-grained multi-class classification of illicit content on specialized, multilingual datasets. Existing comparisons often focus on general NLP benchmarks or different application domains. Secondly, while multilingualism is acknowledged as a challenge, evaluations of LLMs on genuinely multilingual illicit content datasets from ``in-the-wild'' marketplaces are not yet widespread. Finally, understanding the practical trade-offs (performance vs. computational cost) of using PEFT and quantization for these newer LLMs in this specific context is crucial for real-world deployment.

To provide a clear overview of the landscape, this study  includes Table \ref{tab:1} compiling key methodologies, datasets, and performance metrics from prior works, including some papers not discussed in the main text. This aids in contextualizing the current research within the broader field.

This research addresses these gaps by providing a systematic comparison of fine-tuned Llama 3.2 and Gemma 3 models against traditional and transformer baselines using the robust, multilingual DUTA10K dataset. By focusing on these advanced, openly available LLMs, the study aims to comprehensively evaluate their efficacy for nuanced illicit content detection and establish rigorous benchmarks, thereby contributing significantly to the field.

\begin{table*}[hpt]
    \centering
    \scriptsize
    \caption{A summary of related studies in detecting online illicit content}
    \label{tab:1}
    \begin{tabular}{|>{\raggedright\arraybackslash}p{0.06\linewidth}|>{\raggedright\arraybackslash}p{0.16\linewidth}|>{\raggedright\arraybackslash}p{0.27\linewidth}|>{\raggedright\arraybackslash}p{0.22\linewidth}|>{\raggedright\arraybackslash}p{0.15\linewidth}|} % p{0.3\linewidth}
        \toprule
        \textbf{References}& \textbf{Data size} &\textbf{Features} &\textbf{Best model}&\textbf{Result}\\ 
        \midrule
        \cite{RN30}& 6,836 docs& BOW \& TF-IDF & Logistic Regression & 96.60\% Accuracy\\ 
        \hline 
        \cite{sharif2020detecting} &  7,000 docs& BOW \& TF-IDF& Stochastic Gradient Descent (SGD)&84.57\% Accuracy\\
        \hline
        \cite{RN35} & 1,000 docs& TF-IDF, Glasgow \& Entropy& SVM &93.00\% Accuracy\\ \hline 
        \cite{RN49}& 7,997 images& SIFT \& SURF& KNN&82.26\% Accuracy\\ 
        \hline 
        \cite{RN45}& 4,851 docs& TF-IDF& Naive Bayes&93.50\% Accuracy\\ 
        \hline
 \cite{RN32}& 248,359 market listings& Word-based features & MALLET Topic Modeling&81.5\% Precision\\\hline
 \cite{RN18}& 43,607 Instagram comments& Biterm-based features & Biterm Topic Modeling&5,589 comments detected\\\hline
 \cite{RN47}& 2,771,730 tweets& Biterm-based features & Biterm Topic Modeling&1,044,794 tweets detected\\\hline
 \cite{RN25}& 213,041 tweets& Biterm-based features & Biterm Topic Modeling&0.32\% Precision\\\hline
 \cite{RN48}& 141,530 data instances& PCA, tSNE, KPCA& VGG19 + Random Forest + CNN&96\% Accuracy\\\hline
 \cite{RN16}& 6,261 images& CNN deep features & CNN&0.95 F-score\\\hline
 \cite{RN26}& 3,590 images& CNN deep features & CNN&92-99\% Accuracy\\\hline
 \cite{RN41}& 5,500 images& SNN deep features & Siamese Neural Network (SNN)&90.90\% Accuracy\\\hline
 \cite{RN44}& 185,460 market listings& BERT, Doc2Vec for text, ResNet50 for images& SNN&97.54\% Accuracy\\\hline
 \cite{RN6}& 12,857 social media posts& Text and hashtag vectors & LSTM&95\% F1-score\\\hline
 \cite{RN28}& 109,692 market listing& Pre-trained word embedding & LSTM, BERT&96\% Accuracy\\\hline
 \cite{RN21}& 79,705 Instagram posts& BERT for text, VGG19 for images& Meta-Heterogeneous Graph (MetaHG)&93.54\% Accuracy\\\hline
    \end{tabular}

\end{table*}

\section{Methodology}
\label{sec:methodology}

This research employed a systematic methodology to evaluate the efficacy of LLMs for classifying illicit online content, comparing their performance against a set of baseline models. This section details the dataset, preprocessing techniques, experimental setup, model architectures, and evaluation metrics, emphasizing the justification for each methodological choice. The primary LLMs investigated were Meta's Llama 3.2 and Google's Gemma 3. These models were selected due to their representation of current state-of-the-art capabilities from different leading AI research entities, their distinct architectural designs (offering a broader assessment of LLM applicability), their open availability (which supports reproducibility and further research), and their potential for efficient fine-tuning. The baseline models included traditional machine learning approaches, SVM and Multinomial Naive Bayes (MNB), and a prominent transformer-based model, BERT, to provide a robust spectrum for comparison against both classical techniques and a foundational transformer architecture.

\subsection{Research Design Overview}
\label{sec:subsection_research_design}

The study was designed to rigorously assess LLMs in two primary classification tasks. The first is a binary classification task, aimed at distinguishing `illicit' from `non-illicit' content, which assesses the fundamental ability of models to detect the presence of illicit material. The second is a multi-class classification task, focused on categorizing content into 40 specific illicit types (e.g., ``Counterfeit Credit-Cards'', ``Drugs\_Illegal''), which tests the models' capacity for more nuanced, fine-grained understanding crucial for targeted moderation and analysis. This dual-task approach was adopted to provide a comprehensive understanding of model capabilities across different levels of classification granularity. The performance of fine-tuned Llama 3.2 and Gemma 3 models was benchmarked against MNB, SVM, and BERT. MNB and SVM were selected as they are well-established, computationally efficient, and commonly used baselines in text classification, offering a performance reference for traditional methods. BERT (specifically bert-base-uncased), introduced by \cite{RN67}, represents a strong transformer-based baseline. Its inclusion allows for comparison against a model with similar architectural underpinnings (transformers) to the primary LLMs, though typically differing in scale, pre-training objectives, and fine-tuning approaches.

\subsection{Investigated Dataset}
\label{sec:subsection_datasets}

\subsubsection{Dataset Curation and Characteristics}
\label{sec:subsubsection1}

A consolidated JSONL dataset was specifically constructed for this research, derived from raw text files associated with online sources (identified by Onion Addresses from the DUTA10K dataset) and linked with structured metadata from \cite{RN30}. The DUTA10K dataset, provided by Universidad de León, was chosen due to its direct relevance to illicit online content, its multilingual nature, and its public availability for research purposes, facilitating ecologically valid assessments. This dataset exhibits substantial linguistic diversity, an essential characteristic for ecologically valid research in illicit content detection. Comprising entries in over 20 languages, the dataset is predominantly composed of English-language content (approximately 85\%, or 8,785 out of 10,367 entries). However, it also includes significant portions in Russian (508 entries), French (251), German (122), Italian (103), Portuguese (74), and Spanish (73), alongside less common languages such as Danish, Finnish, Catalan, Dutch, Polish, Turkish, Indonesian, and Tagalog. This multilingual nature reflects the global scope of illicit online activity and presents unique challenges for classification tasks. 

\begin{figure}[ht]
  \centering
\includegraphics[width=\linewidth]{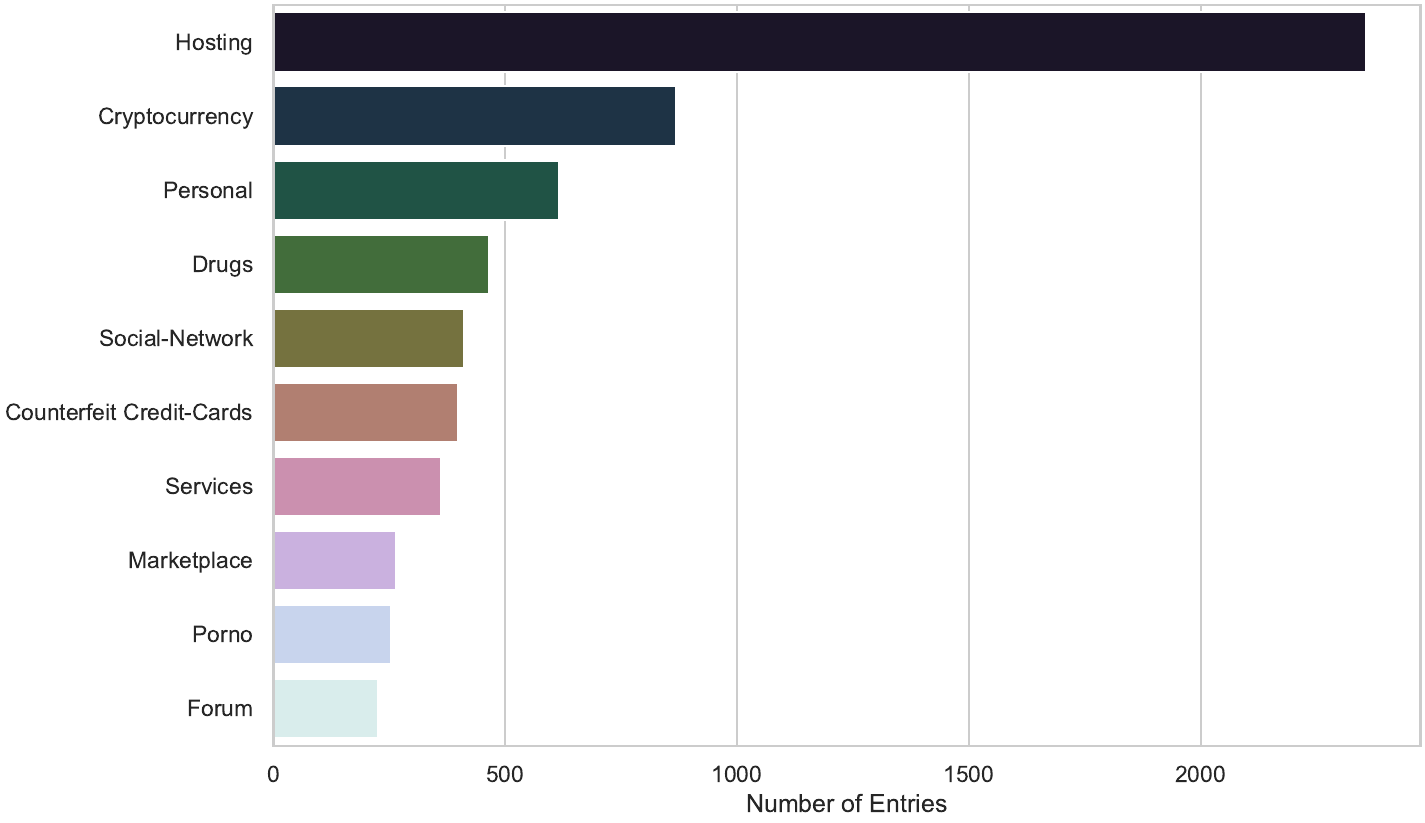}
  \caption{Top 10 Main Categories in the DUTA10K Dataset.}
  \label{fig:1}
\end{figure}

Consequently, preprocessing strategies were carefully designed to accommodate linguistic variation, for instance, through multilingual stopword removal for traditional models and language-agnostic tokenization for transformer-based models. The curated dataset for this study comprises 4,178 instances. Each instance contains the raw textual content from the online source (text), a binary indicator for illicit/non-illicit status (binary\_label) derived directly from the ``illicit'' field in the source metadata, the detected language of the text (lang), the source identifier (onion\_address), and one of 40 specific illicit categories (multiclass\_label). These multi-class categories, as shown in Fig. \ref{fig:1}, were systematically formed by combining ``Main\_Class'' and ``Sub\_Class'' metadata fields, allowing for a granular classification that reflects the diverse nature of illicit activities.

For robust model development and unbiased evaluation, the dataset was consistently split into training (approximately 80\%, 3,342 samples), validation (approximately 10\%, 418 samples), and testing (approximately 10\%, 418 samples) sets. This 80/10/10 split is a standard practice, providing sufficient data for training while reserving distinct sets for hyperparameter tuning (validation) and final performance assessment (testing). For the MNB and SVM baseline models, stratification by label was employed during splitting to ensure that class proportions were maintained across the sets, which is crucial for potentially imbalanced datasets. For transformer-based models (BERT, Llama, Gemma), the data was shuffled before partitioning to ensure random distribution and prevent any ordering bias from affecting the training process.

\subsubsection{Ethical Considerations}
\label{sec:subsubsection2}

The research utilized the DUTA10K dataset, which is publicly available and intended for research purposes related to understanding and combating illicit online activities. All data used in this study consisted of textual content from websites; no personally identifiable information beyond the source Onion Addresses (which are pseudo-anonymous identifiers of services, not individuals) was processed or stored. The textual content was treated as data for natural language processing tasks, with the primary goal of developing models to detect and classify types of illicit content, not to identify or profile individuals or specific service operators. The research aims to contribute to online safety and the development of tools for law enforcement and platform moderation, which is considered an ethical application of AI. The study adheres to responsible research practices by focusing on aggregated content analysis and model performance evaluation. No attempt was made to de-anonymize sources or access live illicit services.

\subsection{Preprocessing}
\label{sec:subsection_preprocessing}
\begin{figure}[ht]
  \centering
  \includegraphics[width=\linewidth]{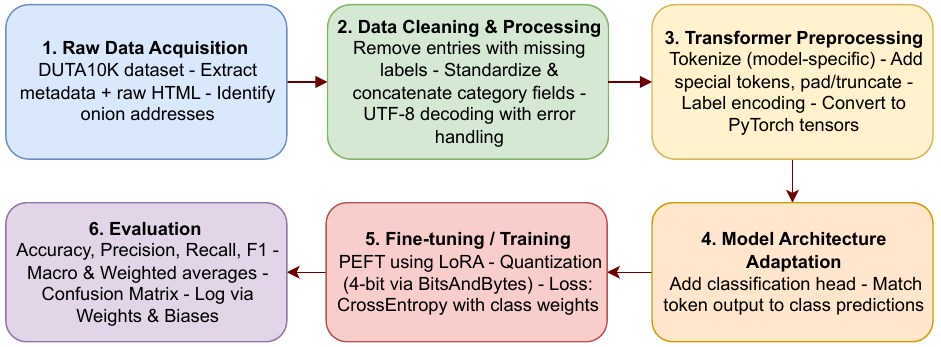}
  \caption{Llama and Gemma training pipeline for text classification.}
  \label{fig:2}
\end{figure}

The data preprocessing pipeline was a critical phase, involving initial dataset construction from raw sources, followed by model-specific preprocessing tailored for the baseline and transformer-based models. These steps were justified by the need to ensure data quality, consistency, and an optimal format for each model type. Fig.~\ref{fig:2} displays the pipeline of model development and training.

Initial dataset construction involved a custom Python script for loading and cleaning the source metadata. This included filtering out entries with missing illicit status information to avoid ambiguous training signals, and standardizing category fields by converting them to string data types and removing extraneous whitespace to ensure uniformity. Unified category labels for multi-class tasks were generated by concatenating main and sub-classification fields. Textual content was extracted from source files using `utf-8' encoding, with robust error handling (e.g., errors=`replace') implemented to manage potential problematic characters or encoding issues common in web-scraped data. This meticulous preparation was fundamental for establishing a high-quality foundation.

For the baseline models (MNB and SVM), a series of standard text normalization techniques were applied. Text was converted to lowercase to treat words like ``Drugs'' and ``drugs'' as identical, reducing vocabulary size and data sparsity. Tokenization was performed using regular expressions to capture alphanumeric sequences as individual words, a common and effective method for text segmentation. Multilingual stop-word removal, leveraging resources like the NLTK library, was facilitated to remove common words (e.g., ``the'', ``is'', ``in'', and their equivalents in other detected languages) that are frequent but often carry little semantic weight for classification; their removal helps focus the model on more informative terms and reduces feature dimensionality, a step adapted for various languages given the dataset's nature. The processed text was then vectorized using the Term Frequency-Inverse Document Frequency (TF-IDF) approach, with the vocabulary limited to the top 5,000 features. TF-IDF was chosen for its proven effectiveness in transforming text into a numerical format suitable for traditional ML algorithms, as it weights terms based on their frequency within a document and rarity across the corpus, highlighting characteristic terms, while feature limitation helps manage computational complexity.

For all transformer-based models (BERT, Llama 3.2, and Gemma 3), preprocessing leveraged their respective model-specific tokenizers, accessed via the Hugging Face Transformers library \cite{RN91}, an essential step as these models are pre-trained with specific tokenization schemes (often subword tokenization like WordPiece or BPE), ensuring compatibility and effective use of learned representations; for BERT, the bert-base-uncased tokenizer was used with the case-lowering feature. The process involved tokenizing input text into subword units, then padding sequences to a uniform maximum length or truncating them if they exceeded this length to ensure consistent input dimensions. Model-specific special tokens (e.g., [CLS], [SEP] for BERT; \texttt{<s>}, \texttt{</s>} for Llama/Gemma) were automatically added. A key step in preparing the Llama and Gemma models for the classification task was the adaptation of their architecture, which involved incorporating a classification head to process the tokenized sequences for this specific objective. 

\begin{figure}[ht]
  \centering
  \includegraphics[width=\linewidth]{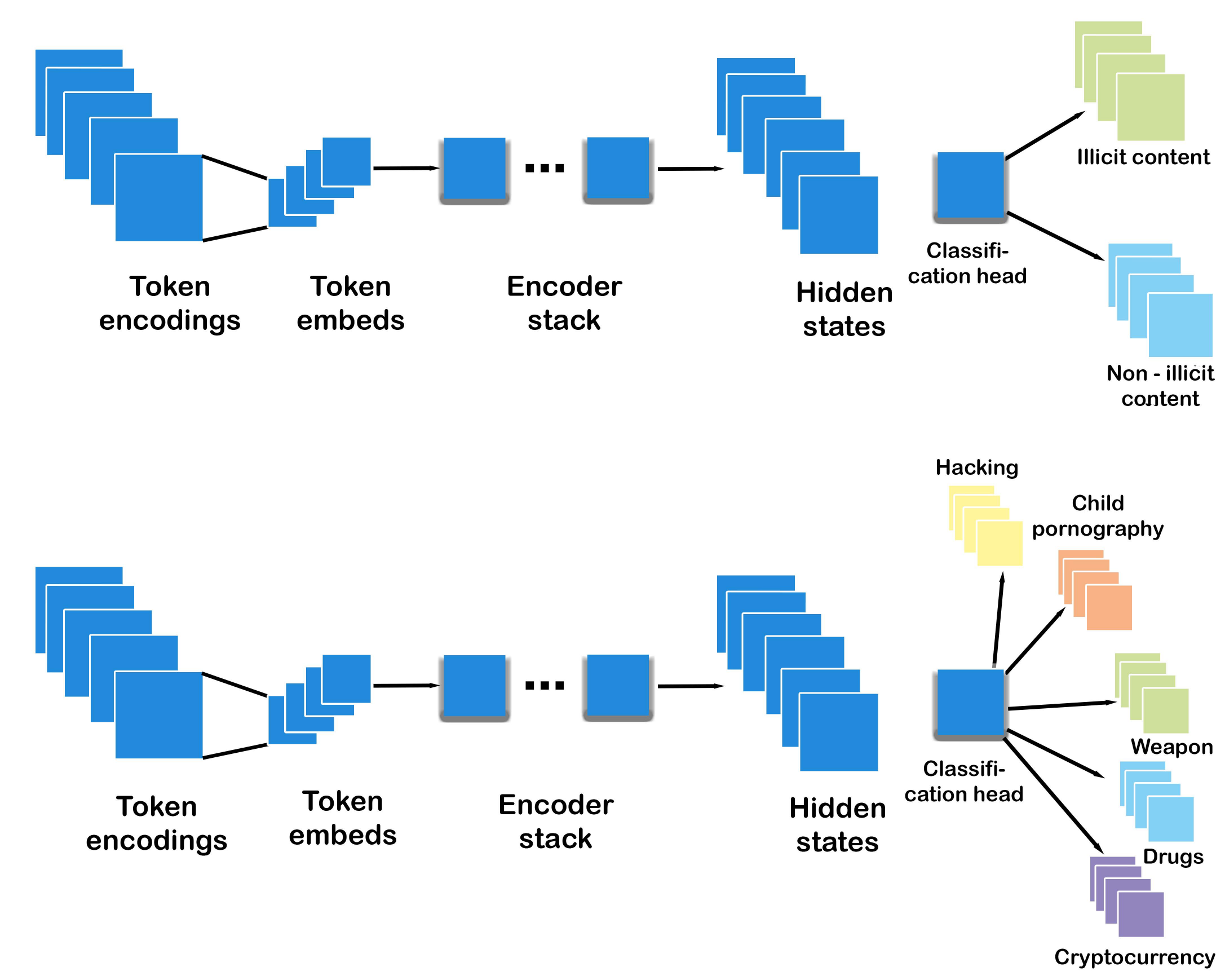}
  \caption{LLM structure in binary and multi-class text classification tasks.}
  \label{fig:3}
\end{figure}

Fig.~\ref{fig:3} shows the customised LLM structures specialised for binary and multi-class tasks. For multi-class tasks, textual category labels were encoded into numerical IDs using Scikit-learn's LabelEncoder or similar techniques. Finally, the tokenized data was converted into PyTorch tensors, the required numerical format.

\subsection{Experimental Setup and Models}
\label{sec:subsection1}

\subsubsection{Environment and Baseline Model Implementation}
\label{sec:subsubsection3}
All experiments were conducted within a Python-based environment. The Hugging Face ecosystem, including Transformers, PEFT, and BitsAndBytes \cite{dettmers2022gpt3} for quantization, was heavily utilized for implementing and fine-tuning BERT, Llama 3.2, and Gemma 3. Scikit-learn was employed for implementing traditional baselines (MNB, SVM) and for calculating evaluation metrics. Text preprocessing for MNB and SVM was supported by NLTK, while Pandas and NumPy handled data manipulation and numerical operations. Weights \& Biases was integrated for systematic experiment tracking, logging configurations, and results to ensure reproducibility. The MNB model followed principles outlined by \cite{RN82}, suitable for text classification with discrete features. The linear SVM was chosen for its effectiveness in high-dimensional spaces. The BERT baseline utilized the bert-base-uncased model, adapted for sequence classification by adding a classification head.

\subsubsection{LLM and BERT Fine-tuning Strategy}
\label{sec:subsubsection4}
The primary LLMs, Llama 3.2 (specifically Llama-3.2-3B) and Gemma 3 (Gemma-3-4B-Instruct), as well as the BERT baseline, were adapted for the classification tasks by adding a sequence classification head atop the pre-trained model. The fine-tuning strategy aimed to balance high performance with computational feasibility, making these powerful models adaptable to specific downstream tasks.

PEFT was implemented using Low-Rank Adaptation (LoRA)~\cite{hu2022lora,nguyen2024fine}, for all three transformer-based models. LoRA was selected because it significantly reduces the number of trainable parameters by injecting smaller, trainable low-rank matrices into specific layers of the transformer architecture, making fine-tuning large models more resource-efficient without sacrificing substantial performance. LoRA configurations involved setting appropriate parameters for rank (r), alpha (lora\_alpha), and target modules. A LoRA dropout of 0.1 was commonly applied to prevent overfitting within the adapter layers. To further optimize resource usage, particularly during binary classification, 4-bit quantization was applied for Llama and Gemma using the BitsAndBytes library, following principles similar to QLoRA \cite{dettmers2022gpt3}, to reduce memory footprint and computational cost. For Gemma's multi-class classification, 4-bit quantization was initially attempted but the model was ultimately loaded without it in some experiments due to technical or performance considerations, operating at bfloat16 precision instead. This choice of quantization was justified by the need to manage the significant computational demands of these large models on available hardware.

All transformer models were fine-tuned using the Hugging Face Trainer API. The training procedure adhered to standard best practices, utilizing the AdamW optimizer, appropriate learning rates (e.g., 2e-5, 5e-5) often with a scheduler, and an effective batch size achieved through per-device settings and gradient accumulation. Models were typically trained for a moderate number of epochs (e.g., around 3-10), with mixed-precision training (fp16=True) enabled where applicable. The best model checkpoint, based on validation performance (e.g., accuracy or weighted F1-score) using CrossEntropyLoss as the criterion, was saved, and early stopping was employed in some configurations to prevent overfitting. For multi-class tasks, the significant class imbalance in the 40 categories was addressed by integrating class weights (inversely proportional to class frequencies) into the CrossEntropyLoss function for Llama 3.2, Gemma 3, and BERT. This technique is justified as it helps prevent model bias towards majority classes and improves performance on minority classes.

\subsection{Evaluation}
\label{sec:subsection2}

Model performance was rigorously assessed using a standard suite of metrics, chosen to provide a holistic view of classifier performance, especially with potential class imbalance. These included Accuracy, measuring the overall proportion of correct classifications; Precision, quantifying the proportion of true positive predictions among all positive predictions and thus the reliability of positive calls; Recall (or Sensitivity), assessing the proportion of true positive predictions among all actual positive instances, indicating the ability to identify all relevant instances; and the F1-score, which is the harmonic mean of precision and recall, offering a balanced measure particularly useful in imbalanced datasets. For multi-class tasks, these metrics were reported as Weighted averages, to account for class imbalance by weighting each class's score by its presence in the dataset, and Macro averages, which calculate the metric independently for each class and then take the unweighted mean, treating all classes equally. The Macro average is particularly insightful for understanding performance on less frequent but potentially critical categories. Scikit-learn's classification\_report and confusion\_matrix functions were utilized for detailed per-class insights and error analysis, with zero\_division=0 specified to handle classes with no predictions without errors. All experimental results, configurations, and metrics were systematically logged using Weights \& Biases for thorough tracking and reproducibility.

\section{Results and Discussions}
\label{sec:results_discussions}

The empirical results from the systematic evaluation of all implemented models are summarized in Table \ref{tab:2} (binary classification) and Table \ref{tab:3} (multi-class classification). These results showcase performance across two distinct classification tasks: binary (illicit vs. non-illicit) and multi-class (40 specific illicit categories). 

\begin{table*}[hpt]
    \centering
    \scriptsize
    \caption{A comparison between LLMs and baseline models in binary classification}
    \label{tab:2}
    \begin{tabular}{|>{\raggedright\arraybackslash}p{0.07\linewidth}|>{\raggedright\arraybackslash}p{0.06\linewidth}|>{\raggedright\arraybackslash}p{0.06\linewidth}|>{\raggedright\arraybackslash}p{0.06\linewidth}|>{\raggedright\arraybackslash}p{0.06\linewidth}|>{\raggedright\arraybackslash}p{0.06\linewidth}|>{\raggedright\arraybackslash}p{0.06\linewidth}|>{\raggedright\arraybackslash}p{0.06\linewidth}|>{\raggedright\arraybackslash}p{0.06\linewidth}|>{\raggedright\arraybackslash}p{0.06\linewidth}|>{\raggedright\arraybackslash}p{0.06\linewidth}|} % p{0.3\linewidth}
        \toprule
        \textbf{Model Type}& \multicolumn{4}{|c|}{\textbf{LLM}}&\multicolumn{6}{|c|}{\textbf{Baseline Model}}\\ 
        \midrule
        Model& \multicolumn{2}{|c|}{Gemma}& \multicolumn{2}{|c|}{Llama}&  \multicolumn{2}{|c|}{Naive Bayes}& \multicolumn{2}{|c|}{SVM}& \multicolumn{2}{|c|}{BERT}\\ 
        \hline 
        Result& Macro& Weighted& Macro&Weighted&  Macro&Weighted& Macro& Weighted& Macro&Weighted\\ \hline 
        Precision&  0.73& 0.81& 0.87&0.89&  0.82&0.85& 0.88& 0.9& 0.92&0.86\\
        \hline
        Recall& 0.68& 0.83& 0.77&0.89&  0.69&0.86& 0.77& 0.9& 0.57&0.84\\ 
        \hline 
        F1& 0.7& 0.82& 0.8&0.88&  0.73&0.85& 0.81& 0.89& 0.58&0.78\\ 
        \bottomrule
 Accuracy& \multicolumn{2}{|c|}{0.83}& \multicolumn{2}{|c|}{0.89}&  \multicolumn{2}{|c|}{0.86}& \multicolumn{2}{|c|}{\textbf{0.9}}& \multicolumn{2}{|c|}{0.84}\\\hline
    \end{tabular}

\end{table*}

\begin{table*}[hpt]
    \centering
    \scriptsize
    \caption{A comparison between LLMs and baseline models in multi-class classification}
    \label{tab:3}
    \begin{tabular}{|>{\raggedright\arraybackslash}p{0.07\linewidth}|>{\raggedright\arraybackslash}p{0.06\linewidth}|>{\raggedright\arraybackslash}p{0.06\linewidth}|>{\raggedright\arraybackslash}p{0.06\linewidth}|>{\raggedright\arraybackslash}p{0.06\linewidth}|>{\raggedright\arraybackslash}p{0.06\linewidth}|>{\raggedright\arraybackslash}p{0.06\linewidth}|>{\raggedright\arraybackslash}p{0.06\linewidth}|>{\raggedright\arraybackslash}p{0.06\linewidth}|>{\raggedright\arraybackslash}p{0.06\linewidth}|>{\raggedright\arraybackslash}p{0.06\linewidth}|} % p{0.3\linewidth}
        \toprule
        \textbf{Model Type}& \multicolumn{4}{|c|}{\textbf{LLM}}&\multicolumn{6}{|c|}{\textbf{Baseline Model}}\\ 
        \midrule
        Model& \multicolumn{2}{|c|}{Gemma}& \multicolumn{2}{|c|}{Llama}&  \multicolumn{2}{|c|}{Naive Bayes}& \multicolumn{2}{|c|}{SVM}& \multicolumn{2}{|c|}{BERT}\\ 
        \hline 
        Result& Macro& Weighted& Macro&Weighted&  Macro&Weighted& Macro& Weighted& Macro&Weighted\\ \hline 
        Precision&  0.62& 0.68& 0.68&0.75&  0.23&0.53& 0.56& 0.78& 0.34&0.6\\
        \hline
        Recall& 0.51& 0.68& 0.58&0.74&  0.16&0.54& 0.4& 0.72& 0.37&0.68\\ 
        \hline 
        F1& 0.54& 0.66& 0.61&0.73&  0.16&0.48& 0.44& 0.72& 0.34&0.63\\ 
        \bottomrule
 Accuracy& \multicolumn{2}{|c|}{0.68}& \multicolumn{2}{|c|}{\textbf{0.74}}&  \multicolumn{2}{|c|}{0.54}& \multicolumn{2}{|c|}{0.72}& \multicolumn{2}{|c|}{0.68}\\\hline
    \end{tabular}

\end{table*}

\subsection{Binary Classification: Illicit vs. Non-Illicit Content}
\label{sec:subsection3}

This task benchmarked the fundamental capability of models to distinguish generally illicit from non-illicit content, often a first-pass filter in moderation systems.

\subsubsection{Baseline Model Performance}
\label{sec:subsubsection5}
The SVM approach demonstrated the strongest baseline performance (accuracy: 0.90, weighted F1: 0.89, macro F1: 0.81), indicating robust, balanced performance. Naive Bayes also yielded commendable results (accuracy: 0.86, weighted F1: 0.85). BERT achieved an accuracy of 0.84 and a weighted F1 of 0.79, with a lower macro F1 of 0.61. The high efficacy of SVM with TF-IDF for text classification is well-established, adept at high-dimensional sparse data and margin maximization, echoed in harmful content detection studies \cite{RN73}. BERT's respectable but lower performance compared to SVM here might suggest TF-IDF captured highly discriminative signals effectively for this binary task, or that standard BERT fine-tuning was less optimal. BERT's lower macro F1 suggests greater class performance disparity, possibly poorer recall for the minority illicit class.

\subsubsection{LLM Performance}
\label{sec:subsubsection6}
Llama 3.2 achieved an accuracy of 0.89 and a weighted F1-score of 0.88, closely mirroring SVM, with a strong macro F1 of 0.80, superior to BERT's. Gemma 3 scored slightly lower (accuracy: 0.83, weighted F1: 0.82, macro F1: 0.70). Llama 3.2's strong showing can be attributed to its extensive pre-training \cite{RN84} enabling sophisticated language understanding. Differences between Llama 3.2 and Gemma 3 may stem from pre-training data, model scale, or fine-tuning dynamics, with Gemma models aiming for a balance of performance and efficiency~\cite{RN52}. 

\subsubsection{Discussion of Binary Classification Results}
\label{sec:subsubsection7}
For binary classification, the SVM baseline with TF-IDF proved exceptionally competitive, marginally outperforming Llama~3.2. This suggests that for well-defined binary tasks, simpler, less computationally expensive models can achieve excellent results with effective feature engineering, aligning with findings where traditional methods remain potent \cite{RN110}. However, Llama 3.2's near-parity underscores LLMs' capabilities, potentially advantageous where feature engineering is complex or deeper semantic understanding is paramount. Practical choices might weigh deployment cost, inference speed, and interpretability. Gemma 3, while outperforming BERT, did not match Llama~3.2 or SVM, highlighting performance variability among LLMs.

\subsection{Multi-class Classification: Categorizing Illicit Content}
\label{sec:subsection_multiclass}

Categorizing content into 40 illicit types presented a significantly greater challenge due to increased class granularity, semantic overlap, and dataset imbalance, demanding higher model sophistication.

\subsubsection{Baseline Model Performance}
\label{sec:subsubsection8}
SVM again outperformed Naive Bayes among traditional baselines (accuracy: 0.72, weighted F1: 0.72), but its macro F1 dropped to 0.44, indicating struggles with less frequent or harder-to-distinguish categories. Naive Bayes performed poorly (weighted F1: 0.48, macro F1: 0.16). BERT achieved an accuracy of 0.68 and a weighted F1 of 0.63, but its macro F1 of 0.34 was also low. This suggests even transformer models like BERT face challenges with high-cardinality, imbalanced multi-class tasks using standard fine-tuning, despite class weighting, a difficulty noted elsewhere.

\subsubsection{Llama and Gemma Performance}
\label{sec:subsubsection9}
Here, Llama 3.2 and Gemma 3 showcased their advanced capabilities. Llama 3.2 was the top performer (accuracy: 0.74, weighted F1: 0.73). Its macro F1-score of 0.61 was notably better than all baselines, suggesting more balanced performance across categories, likely aided by class weighting. Gemma also performed robustly (accuracy: 0.68, weighted F1: 0.66, macro F1: 0.54), also benefiting from class weighting. Both Llama 3.2 and Gemma 3 significantly outperformed all baselines, including BERT. This underscores the value of their deeper semantic understanding and extensive pre-training for complex classification, where their ability to learn rich representations directly from text is key, potentially reflecting emergent reasoning capabilities \cite{RN112}.

\subsubsection{Discussion of Multi-class Classification Results}
\label{sec:subsubsection10}
The multi-class results affirm the advantages of larger LLMs like Llama and Gemma for nuanced tasks. While all models saw performance decrease from the binary task, Llama 3.2 demonstrated the most robustness. Class weighting was critical for Llama and Gemma in mitigating imbalance and improving performance on less frequent categories. Baselines, including BERT, showed more pronounced drops in macro F1-scores, indicating significant struggles with the high-cardinality, imbalanced nature of the task. The superior performance of Llama and Gemma highlights their advanced representation learning. However, the persistent gap between weighted and macro F1 scores across all models indicates ongoing difficulty with extreme minority classes, suggesting this remains an active and challenging research area.

\subsection{Comparative Analysis and Overall Insights}
\label{sec:subsection_comparative}

\subsubsection{LLMs vs. Baseline Models}
\label{sec:subsubsection11}
The study reveals a task-dependent model hierarchy. For simpler binary classification, SVM was highly competitive with Llama 3.2, both outperforming Gemma 3 and BERT. This suggests that for certain well-defined detection tasks, simpler models can be exceptionally effective and resource-efficient, cautioning against assuming universal superiority of larger models. However, as task complexity increased to 40-class categorization, Llama 3.2 and Gemma 3 demonstrated a clear performance advantage over all baselines, including BERT. This shift indicates that the superior semantic representation capabilities of these very large LLMs become critical for finer-grained distinctions, consistent with trends where larger pre-trained models excel at tasks requiring deeper linguistic insight~\cite{brown2020language}.

\subsubsection{Llama 3.2 vs. Gemma}
\label{sec:subsubsection12}
Llama 3.2 consistently outperformed Gemma 3 across both tasks (binary weighted F1: 0.88 vs 0.82; multi-class weighted F1: 0.73 vs 0.66). This could be attributed to differences in pre-training data composition, architectural scale, and fine-tuning adaptability for this specific domain and dataset. Llama 3.2's higher scores suggest it captured more relevant linguistic features or generalized more effectively.

\subsubsection{Computational Considerations and Resource Efficiency}
\label{sec:subsubsection13}
A crucial practical aspect is the differing computational resources required. Traditional models (SVM, Naive Bayes) are the most computationally efficient. BERT represents an intermediate level. Larger LLMs like Llama 3.2 and Gemma 3, even when fine-tuned using PEFT and quantization, demand significantly more computational power and specialized expertise. SVM's strong binary performance makes it an attractive, efficient option where resources are constrained. This trade-off between performance and cost (computational, financial, and environmental, as noted by \cite{RN108}) is a key practical consideration in deploying such models.

\subsection{Implications of Findings}
\label{sec:subsection_implications}

The findings of this research carry several important implications for the automated detection and classification of illicit online content, particularly regarding the strategic application of Llama and Gemma. A primary implication is the reinforcement of task-dependent model selection; while traditional models like SVM demonstrate high efficacy and efficiency for broader binary detection tasks, the nuanced understanding required for fine-grained multi-class classification benefits significantly from the advanced capabilities of larger LLMs such as Llama 3.2 and Gemma 3, suggesting that a one-size-fits-all approach is suboptimal and that the complexity of the classification problem should guide model choice, balancing performance needs with computational and deployment costs. 

Furthermore, this study affirms the substantial potential of modern LLMs like Llama and Gemma for tackling complex semantic challenges in content moderation, as their ability to learn rich contextual representations directly from text without extensive manual feature engineering is a critical advantage when dealing with the diverse and evolving language of illicit online content, positioning these LLMs as powerful tools for developing more sophisticated and adaptable moderation systems capable of identifying a wider spectrum of harmful material with greater accuracy~\cite{howard2018universal}. 

The research also highlights the criticality of employing strategies to handle data imbalance, such as class weighting, when working with advanced models in multi-class settings, as the observed improvement in macro F1-scores for Llama and Gemma underscores that sophisticated architectures alone are insufficient and must be paired with data-aware training techniques to ensure robust and equitable performance across all categories, especially for detecting less frequent but potentially highly harmful content types. 

Finally, this work contributes to the ongoing benchmarking and understanding of the evolving NLP landscape by comparing Llama and Gemma not only against traditional methods but also against a foundational transformer like BERT, providing insights into the incremental benefits and specific strengths of these newer, larger architectures, particularly their enhanced capacity for fine-grained reasoning and classification in complex domains.

\subsection{Limitations of the Study}
\label{sec:subsection_limitations}

While this research offers valuable insights into the comparative performance of Llama, Gemma, and baseline models for illicit content classification, certain limitations inherent in its scope and methodology must be acknowledged. A primary limitation is the dataset specificity and its implications for generalizability, as conclusions are drawn from experiments on the single, specifically curated DUTA10K dataset; the unique characteristics of this dataset, including its linguistic diversity, the specific distribution and nature of the 40 illicit content categories, and the prevalent writing styles, may not fully represent the vast spectrum of illicit online environments, and consequently, model performance might differ on other datasets from varied sources or with different content characteristics. 

The study also focused on specific accessible variants and scales of the LLMs (e.g., conceptual 3B for Llama 3.2, 4B for Gemma 3) and a base version of BERT; it is plausible that different versions or scales of Llama and Gemma, or LLMs with more extensive pre-training on relevant corpora (e.g., security or legal texts), could yield different performance outcomes, and similarly, the PEFT configurations, while based on common practices, might not be globally optimal for this specific dataset and tasks. 

Another significant consideration is the dynamic and adversarial nature of illicit content, as those engaged in illicit activities constantly adapt their language, employ obfuscation, and evolve their methods to evade detection \cite{RN93}; models trained on a static dataset, including sophisticated LLMs like Llama and Gemma \cite{nguyen2025large}, are susceptible to performance degradation over time as new patterns emerge (concept drift), and this study did not explicitly evaluate adversarial robustness or the models' capacity for continuous adaptation. 

Computational resources also present a practical limitation; while PEFT and quantization were used to make fine-tuning Llama and Gemma more manageable, these state-of-the-art LLMs still demand considerable computational power compared to the baseline models, a factor that can influence their feasibility for deployment in all operational contexts, especially for organizations with limited resources. 

Furthermore, the interpretability of advanced models like Llama and Gemma remains a challenge, as their ``black box'' nature makes it difficult to understand the precise reasoning behind their classifications, which is a significant concern in sensitive applications like content moderation where accountability and bias detection are paramount \cite{RN105}. 

\section{Conclusions and Future Work}
\label{sec:conclusions}

This study provides a comprehensive comparative analysis of LLMs, Llama 3.2 and Gemma 3, against diverse baselines for illicit content classification. The results reveal a nuanced, task-dependent model efficacy. While traditional models like SVMs demonstrate exceptional performance and efficiency for binary detection, larger LLMs, particularly Llama 3.2, exhibit a significant advantage in complex, multi-class categorization demanding deep semantic understanding. Gemma 3 also showed strong promise in these multi-class settings. The choice of model in practical applications must be guided by task requirements, resource availability, and trade-offs between performance, efficiency, and interpretability. Strategic application of PEFT, quantization, and methods to address class imbalance are crucial for harnessing advanced LLMs. Despite limitations, this research underscores the substantial potential of modern LLMs like Llama and Gemma to significantly advance automated systems combating illicit online content.

Building upon these insights, several promising avenues for future research emerge. Investigations could explore even larger-scale LLMs, models with domain-specific pre-training, or those leveraging Mixture-of-Experts (MoE) architectures, which might allow for specialized `experts' within the model to focus on different illicit vernaculars or categories, potentially improving multi-class performance. Concurrently, refining PEFT methods and prompt engineering \cite{RN112} remains vital for efficient adaptation, especially with limited data.

The development of hybrid architectures remains a compelling direction, combining efficient baselines with Llama or Gemma for nuanced analysis. Expanding the scope to address robust cross-lingual and multimodal (text, image, video) illicit content detection is crucial \cite{RN66}, as illicit activities increasingly span languages and media formats; extending models like Llama and Gemma to handle these inputs is a key next step.

Because online threats are always changing and can be hostile \cite{RN93}, it is very important to establish frameworks for ongoing learning and adaptation. This includes exploring the potential of AI agents for proactive detection while developing countermeasures against their malicious use, and enhancing adversarial robustness to combat obfuscation techniques. Furthermore, enhancing interpretability and explainability (XAI)~\cite{patel2025misinformation} is critical for building trust, identifying biases, and facilitating human oversight. Integrating Constitutional AI principles and guardrails during training and deployment can help ensure these powerful models operate within ethical and legal boundaries, promoting responsible AI use in content moderation.

Addressing data handling concerns through privacy-preserving AI techniques, such as federated learning or differential privacy, will be important when training on sensitive datasets. Finally, fine-grained error analysis focusing on underrepresented categories can inform targeted strategies like advanced data augmentation \cite{RN117} or specialized loss functions. Embracing these future directions, continued LLM advancements hold considerable promise for breakthroughs in online safety.

\bibliographystyle{IEEEtran}
\balance
\bibliography{cas-refs}

@article{sharif2020detecting,
  title={Detecting suspicious texts using machine learning techniques},
  author={Sharif, Omar and Hoque, Mohammed Moshiul and Kayes, ASM and Nowrozy, Raza and Sarker, Iqbal H},
  journal={Applied Sciences},
  volume={10},
  number={18},
  pages={6527},
  year={2020},
  publisher={MDPI}
}

@inproceedings{nguyen2024fine,
  title={Fine-tuning {Llama} 2 large language models for detecting online sexual predatory chats and abusive texts},
  author={Nguyen, Thanh Thi and Wilson, Campbell and Dalins, Janis},
  booktitle={Proceedings of the 32nd European Symposium on Artificial Neural Networks, Computational Intelligence and Machine Learning},
pages={613--618},
  year={2024}
}

@inproceedings{nguyen2025large,
  title={Large Language Models for Detection of Life-Threatening Texts},
  author={Nguyen, Thanh Thi and Wilson, Campbell and Dalins, Janis},
  booktitle={Pacific-Asia Conference on Knowledge Discovery and Data Mining},
  pages={311--323},
  year={2025},
  organization={Springer}
}

@article{patel2025misinformation,
  title={Misinformation Detection using Large Language Models with Explainability},
  author={Patel, Jainee and Bhatt, Chintan and Trivedi, Himani and Nguyen, Thanh Thi},
  journal={arXiv preprint arXiv:2510.18918},
  year={2025}
}

@article{RN47,
   author = {Mackey, Tim K. and Purushothaman, Vidya and Haupt, Michael and Nali, Matthew C. and Li, Jiawei},
   title = {Application of unsupervised machine learning to identify and characterise hydroxychloroquine misinformation on Twitter},
   journal = {The Lancet Digital Health},
   volume = {3},
   number = {2},
   pages = {e72-e75},
   year = {2021},
   type = {Journal Article}
}

@inproceedings{RN110,
   author = {Wang, Sida and Manning, Christopher},
   title = {Baselines and Bigrams: Simple, Good Sentiment and Topic Classification},
   booktitle = {Proceedings of the 50th Annual Meeting of the ACL}, year = {2012},
   pages = {90-94},
}

@inproceedings{RN67,
   author = {Devlin, Jacob and Chang, Ming-Wei and Lee, Kenton and Toutanova, Kristina},
   title = {{BERT}: Pre-training of Deep Bidirectional Transformers for Language Understanding},
   booktitle = {Proceedings of the 2019 Conference of the North American Chapter of the ACL: Human Language Technologies},
   pages = {4171-4186}, year = {2019},
   DOI = {10.18653/v1/N19-1423},
}

@article{RN112,
  title={Chain-of-thought prompting elicits reasoning in large language models},
  author={Wei, Jason and Wang, Xuezhi and Schuurmans, Dale and Bosma, Maarten and Xia, Fei and Chi, Ed and Le, Quoc V and Zhou, Denny and others},
  journal={Advances in Neural Information Processing Systems},
  volume={35},
  pages={24824--24837},
  year={2022}
}

@inproceedings{RN45,
  title={Classification of illegal activities on the dark web},
  author={He, Siyu and He, Yongzhong and Li, Mingzhe},
  booktitle={Proceedings of the 2nd International Conference on Information Science and Systems},
  pages={73--78},
  year={2019}
}

@inproceedings{RN30,
   author = {Al Nabki, Wesam and Fidalgo, Eduardo and Alegre, Enrique and Paz, Ivan},
   title = {Classifying Illegal Activities on Tor Network Based on Web Textual Contents},
   booktitle = {Proceedings of the 15th Conference of the European Chapter of the ACL},
   pages = {35-43}, year =  {2017},
   DOI = {10.18653/v1/E17-1004},
}

@inproceedings{RN49,
  title={Comparison of {SIFT} and {SURF} Methods for Porn Image Detection},
  author={Setyanto, Arief and Kusrini, Kusrini and Agastya, I Made Artha and others},
  booktitle={4th International Conference on Information Technology, Information Systems and Electrical Engineering (ICITISEE)},
  pages={281--285},
  year={2019},
  organization={IEEE}
}

@article{RN11,
   author = {Yenala, Harish and Jhanwar, Ashish and Chinnakotla, Manoj K. and Goyal, Jay},
   title = {Deep learning for detecting inappropriate content in text},
   journal = {International Journal of Data Science and Analytics},
   volume = {6},
   number = {4},
   pages = {273-286},
   ISSN = {2364-4168},
   DOI = {10.1007/s41060-017-0088-4},
   year = {2018},
   type = {Journal Article}
}

@article{RN48,
   author = {Singh, Deepak and Shukla, Anurag and Sajwan, Mohit},
   title = {Deep transfer learning framework for the identification of malicious activities to combat cyberattack},
   journal = {Future Generation Computer Systems},
   volume = {125},
   pages = {687-697},
   ISSN = {0167-739X},
   year = {2021},
   type = {Journal Article}
}

@article{RN32,
   author = {Li, Zhengyi and Du, Xiangyu and Liao, Xiaojing and Jiang, Xiaoqian and Champagne-Langabeer, Tiffany},
   title = {Demystifying the Dark Web Opioid Trade: Content Analysis on Anonymous Market Listings and Forum Posts},
   journal = {Journal of Medical Internet Research},
   volume = {23},
   number = {2},
   year = {2021},
   type = {Journal Article}
}

@article{RN17,
   author = {Arora, Arnav and Nakov, Preslav and Hardalov, Momchil and Sarwar, Sheikh Muhammad and Nayak, Vibha and Dinkov, Yoan and Zlatkova, Dimitrina and Dent, Kyle and Bhatawdekar, Ameya and Bouchard, Guillaume and Augenstein, Isabelle},
   title = {Detecting Harmful Content on Online Platforms: What Platforms Need vs. Where Research Efforts Go},
   journal = {ACM Computing Surveys},
   volume = {56},
   DOI = {10.1145/3603399},
   year = {2023},
   type = {Journal Article}
}

@article{RN21,
  title={Distilling meta knowledge on heterogeneous graph for illicit drug trafficker detection on social media},
  author={Qian, Yiyue and Zhang, Yiming and Ye, Yanfang and Zhang, Chuxu and others},
  journal={Advances in Neural Information Processing Systems},
  volume={34},
  pages={26911--26923},
  year={2021}
}

@inproceedings{RN108,
  title={Energy and policy considerations for modern deep learning research},
  author={Strubell, Emma and Ganesh, Ananya and McCallum, Andrew},
  booktitle={Proceedings of the AAAI Conference on Artificial Intelligence},
  volume={34},
  number={09},
  pages={13693--13696},
  year={2020}
}

@article{RN41,
   author = {Cascavilla, Giuseppe and Catolino, Gemma and Conti, Mauro and Mellios, Dimos and Tamburri, Damian},
   title = {Few Images, Many Insights: Illicit Content Detection Using a Limited Number of Images},
   journal = {ACM Transactions on Intelligent Systems and Technology},
   ISSN = {2157-6904},
   DOI = {10.1145/3696458},
   url = {https://doi.org/10.1145/3696458},
   year = {2024},
   type = {Journal Article}
}

@article{RN20,
   author = {Hu, Chuanbo and Liu, Bin and Ye, Yanfang and Li, Xin},
   title = {Fine-grained classification of drug trafficking based on Instagram hashtags},
   journal = {Decision Support Systems},
   volume = {165},
   pages = {113896},
   ISSN = {0167-9236},
   DOI = {https://doi.org/10.1016/j.dss.2022.113896},
   year = {2023},
   type = {Journal Article}
}

@misc{RN52,
   author = {Hamzah, Shaik},
   title = {Google’s {Gemma} 3: Features, Benchmarks, Performance and Implementation},
   year = {2025},
   url = {https://www.analyticsvidhya.com/blog/2025/03/gemma-3/},
   type = {Blog}
}

@article{RN35,
   author = {Sabbah, Thabit and Selamat, Ali and Selamat, Md Hafiz and Ibrahim, Roliana and Fujita, Hamido},
   title = {Hybridized term-weighting method for Dark Web classification},
   journal = {Neurocomputing},
   volume = {173},
   pages = {1908-1926},
   ISSN = {0925-2312},
   DOI = {https://doi.org/10.1016/j.neucom.2015.09.063},
   year = {2016},
   type = {Journal Article}
}

@inproceedings{RN29,
   author = {Shah, D. and Harrison, T. G. and Freas, C. B. and Maimon, D. and Harrison, R. W.},
   title = {Illicit Activity Detection in Large-Scale Dark and Opaque Web Social Networks},
   booktitle = {IEEE International Conference on Big Data},
   pages = {4341-4350}, year = {2020},
   DOI = {10.1109/BigData50022.2020.9378229},
}

@inproceedings{RN51,
   author = {Cascavilla, Giuseppe and Catolino, Gemma and Sangiovanni, Mirella},
   title = {Illicit Darkweb Classification via Natural-language Processing: Classifying Illicit Content of Webpages based on Textual Information},
   booktitle = {19th International Conference on Security and Cryptography, SECRYPT},
   pages = {620--626}, year = {2022},
   ISBN = {9789897585906},
   DOI = {10.5220/0011298600003283},
}

@techreport{RN55,
   author = {Sartor, Giovanni and Loreggia, Andrea},
   title = {The impact of algorithms for online content filtering or moderation: Upload filters},
   url = {europarl.europa.eu/thinktank/en/document/IPOL\_STU(2020)657101},
   year = {2020},
   type = {Report},
institution ={European Parliament Think Tank}
}

@inproceedings{RN73,
   author = {Dadvar, Maral and Trieschnigg, Dolf and Ordelman, Roeland and de Jong, Franciska},
   title = {Improving Cyberbullying Detection with User Context},
   booktitle = {Advances in Information Retrieval},
   pages = {693-696}, year = {2013},
   ISBN = {978-3-642-36973-5},
}

@misc{RN84,
   author = {Meta},
   title = {Introducing {Meta} {Llama} 3: The most capable openly available {LLM} to date},
   volume = {2025},
   url = {https://ai.meta.com/blog/meta-llama-3/},
   year = {2024},
   type = {Blog}
}

@book{RN82,
   author = {Schütze, Hinrich and Manning, Christopher D and Raghavan, Prabhakar},
   title = {Introduction to Information Retrieval},
   publisher = {Cambridge University Press Cambridge},
   volume = {39},
   year = {2008},
   type = {Book}
}

@article{brown2020language,
  title={Language models are few-shot learners},
  author={Brown, Tom and Mann, Benjamin and Ryder, Nick and Subbiah, Melanie and Kaplan, Jared D and Dhariwal, Prafulla and Neelakantan, Arvind and Shyam, Pranav and Sastry, Girish and Askell, Amanda and others},
  journal={Advances in Neural Information Processing Systems},
  volume={33},
  pages={1877--1901},
  year={2020}
}

@inproceedings{RN93,
   author = {Chen, Zhiyuan and Ma, Nianzu and Liu, Bing},
   title = {Lifelong Learning for Sentiment Classification},
   booktitle = {Proceedings of the 53rd Annual Meeting of the ACL and the 7th International Joint Conference on Natural Language Processing},
   pages = {750-756}, year = {2015},
   DOI = {10.3115/v1/P15-2123},
}

@article{dettmers2022gpt3,
  title={{LLM}.int8(): 8-bit matrix multiplication for transformers at scale},
  author={Dettmers, Tim and Lewis, Mike and Belkada, Younes and Zettlemoyer, Luke},
  journal={Advances in Neural Information Processing Systems},
  volume={35},
  pages={30318--30332},
  year={2022}
}

@inproceedings{
hu2022lora,
title={Lo{RA}: Low-Rank Adaptation of Large Language Models},
author={Edward J Hu and yelong shen and Phillip Wallis and Zeyuan Allen-Zhu and Yuanzhi Li and Shean Wang and Lu Wang and Weizhu Chen},
booktitle={International Conference on Learning Representations},
year={2022},
url={https://openreview.net/forum?id=nZeVKeeFYf9}
}

@article{RN6,
   author = {Li, Jiawei and Xu, Qing and Shah, Neal and Mackey, Tim K.},
   title = {A Machine Learning Approach for the Detection and Characterization of Illicit Drug Dealers on Instagram: Model Evaluation Study},
   journal = {Journal of Medical Internet Research},
   volume = {21},
   number = {6},
   pages = {e13803},
   ISSN = {1438-8871},
   DOI = {10.2196/13803},
   year = {2019},
   type = {Journal Article}
}

@article{RN114,
   author = {Munksgaard, Rasmus},
   title = {Marketness and Governance: A Typology of Illicit Online Markets},
   journal = {Deviant Behavior},
   volume = {45},
   number = {12},
   pages = {1711-1728},
   ISSN = {0163-9625},
   year = {2024},
   type = {Journal Article}
}

@inproceedings{guo2024moderating,
  title={Moderating Illicit Online Image Promotion for Unsafe User Generated Content Games Using Large Vision-Language Models},
  author={Guo, Keyan and Utkarsh, Ayush and Ding, Wenbo and Ondracek, Isabelle and Zhao, Ziming and Freeman, Guo and Vishwamitra, Nishant and Hu, Hongxin},
  booktitle={33rd USENIX Security Symposium},
  pages={5787--5804},
  year={2024}
}

@article{RN44,
   author = {Wang, Yilei and Hu, Yuelin and Xu, Wenliang and Zou, Futai},
   title = {Multi-Identity Recognition of Darknet Vendors Based on Metric Learning},
   journal = {Applied Sciences},
   volume = {14},
   number = {4},
   pages = {1619},
   ISSN = {2076-3417},
   DOI = {10.3390/app14041619},
   year = {2024},
   type = {Journal Article}
}

@article{RN53,
   author = {Haynie, Dana L. and Duxbury, Scott W.},
   title = {Online Illegal Cryptomarkets},
   journal = {Annual Review of Sociology},
   volume = {50},
   number = {Volume 50, 2024},
   pages = {671-690},
   ISSN = {1545-2115},
   DOI = {https://doi.org/10.1146/annurev-soc-090523-052916},
   year = {2024},
   type = {Journal Article}
}

@article{RN26,
   author = {Malinverni, Eva Savina and Abate, Dante and Agapiou, Antonia and Stefano, Francesco Di and Felicetti, Andrea and Paolanti, Marina and Pierdicca, Roberto and Zingaretti, Primo},
   title = {SIGNIFICANCE deep learning based platform to fight illicit trafficking of Cultural Heritage goods},
   journal = {Scientific Reports},
   volume = {14},
   number = {1},
   pages = {15081-12},
   ISSN = {2045-2322},
   DOI = {10.1038/s41598-024-65885-6},
   year = {2024},
   type = {Journal Article}
}

@article{RN25,
   author = {Mackey, Tim and Kalyanam, Janani and Klugman, Josh and Kuzmenko, Ella and Gupta, Rashmi},
   title = {Solution to Detect, Classify, and Report Illicit Online Marketing and Sales of Controlled Substances via Twitter: Using Machine Learning and Web Forensics to Combat Digital Opioid Access},
   journal = {Journal of Medical Internet Research},
   volume = {20},
   number = {4},
   pages = {e10029},
   ISSN = {1438-8871},
   DOI = {10.2196/10029},
   year = {2018},
   type = {Journal Article}
}

@article{RN65,
   author = {Al-Nabki, Mhd Wesam and Fidalgo, Eduardo and Alegre, Enrique and Fernández-Robles, Laura},
   title = {{ToRank}: Identifying the most influential suspicious domains in the Tor network},
   journal = {Expert Systems with Applications},
   volume = {123},
   pages = {212-226},
   ISSN = {0957-4174},
   year = {2019},
   type = {Journal Article}
}

@article{RN16,
   author = {Kulkarni, Ritwik and Di Minin, Enrico},
   title = {Towards automatic detection of wildlife trade using machine vision models},
   journal = {Biological Conservation},
   volume = {279},
   pages = {109924},
   ISSN = {0006-3207},
   DOI = {https://doi.org/10.1016/j.biocon.2023.109924},
   year = {2023},
   type = {Journal Article}
}

@article{RN28,
   author = {Kanti Singh, Sangher and Singh, Archana and Pandey, Hari Mohan and Kumar, Vivek},
   title = {Towards Safe Cyber Practices: Developing a Proactive Cyber-Threat Intelligence System for Dark Web Forum Content by Identifying Cybercrimes},
   journal = {Information},
   volume = {14},
   number = {6},
   pages = {349},
   DOI = {https://doi.org/10.3390/info14060349},
   year = {2023},
   type = {Journal Article}
}

@inproceedings{RN91,
   author = {Wolf, Thomas and Debut, Lysandre and Sanh, Victor and Chaumond, Julien and Delangue, Clement and Moi, Anthony and Cistac, Pierric and Rault, Tim and Louf, Remi and Funtowicz, Morgan and Davison, Joe and Shleifer, Sam and von Platen, Patrick and Ma, Clara and Jernite, Yacine and Plu, Julien and Xu, Canwen and Le Scao, Teven and Gugger, Sylvain and Drame, Mariama and Lhoest, Quentin and Rush, Alexander},
   title = {Transformers: State-of-the-Art Natural Language Processing},
   booktitle = {Proceedings of the 2020 Conference on Empirical Methods in Natural Language Processing: System Demonstrations},
   pages = {38-45}, year = {2020},
   DOI = {10.18653/v1/2020.emnlp-demos.6},
}

@inbook{RN34,
   author = {Soldner, Felix and Kleinberg, Bennett and Johnson, Shane},
   title = {Trends in online consumer fraud: A data science perspective},
   booktitle = {A Fresh Look at Fraud},
   publisher = {Routledge},
   address = {United Kingdom},
   volume = {1},
   edition = {1},
   pages = {167-191},
   ISBN = {0367861445},
   DOI = {10.4324/9781003017189-9},
   year = {2022},
   type = {Book Section}
}

@article{RN19,
   author = {Fuller, Ashly and Vasek, Marie and Mariconti, Enrico and Johnson, Shane D.},
   title = {Understanding and preventing the advertisement and sale of illicit drugs to young people through social media: A multidisciplinary scoping review},
   journal = {Drug and Alcohol Review},
   volume = {43},
   number = {1},
   pages = {56-74},
   ISSN = {0959-5236},
   DOI = {https://doi.org/10.1111/dar.13716},
   year = {2024},
   type = {Journal Article}
}

@inproceedings{howard2018universal,
  title={Universal Language Model Fine-tuning for Text Classification},
  author={Howard, Jeremy and Ruder, Sebastian},
  booktitle={Proceedings of the 56th Annual Meeting of the Association for Computational Linguistics},
  pages={328--339},
  year={2018}
}

@inproceedings{RN66,
   author = {Conneau, Alexis and Khandelwal, Kartikay and Goyal, Naman and Chaudhary, Vishrav and Wenzek, Guillaume and Guzmán, Francisco and Grave, Edouard and Ott, Myle and Zettlemoyer, Luke and Stoyanov, Veselin},
   title = {Unsupervised Cross-lingual Representation Learning at Scale},
   booktitle = {Proceedings of the 58th Annual Meeting of the ACL},
   pages = {8440-8451}, year = {2020},
   DOI = {10.18653/v1/2020.acl-main.747},
}

@article{RN18,
   author = {Shah, Neal and Li, Jiawei and Mackey, Tim K.},
   title = {An Unsupervised Machine Learning Approach for the Detection and Characterization of Illicit Drug-Dealing Comments and Interactions on Instagram},
   journal = {Substance Abuse},
   volume = {43},
   number = {1},
   pages = {273-277},
   DOI = {10.1080/08897077.2021.1941508},
   year = {2022},
   type = {Journal Article}
}

@article{RN36,
   author = {Nazah, S. and Huda, S. and Abawajy, J. H. and Hassan, M. M.},
   title = {An Unsupervised Model for Identifying and Characterizing Dark Web Forums},
   journal = {IEEE Access},
   volume = {9},
   pages = {112871-112892},
   ISSN = {2169-3536},
   DOI = {10.1109/ACCESS.2021.3103319},
   year = {2021},
   type = {Journal Article}
}

@article{RN15,
  title={Unveiling the potential of knowledge-prompted {ChatGPT} for enhancing drug trafficking detection on social media},
  author={Hu, Chuanbo and Liu, Bin and Li, Xin and Ye, Yanfang},
  journal={arXiv preprint arXiv:2307.03699},
  year={2023}
}

@article{RN117,
   author = {Sauber-Cole, Rick and Khoshgoftaar, Taghi M.},
   title = {The use of generative adversarial networks to alleviate class imbalance in tabular data: a survey},
   journal = {Journal of Big Data},
   volume = {9},
   number = {1},
   pages = {98},
   ISSN = {2196-1115},
   DOI = {10.1186/s40537-022-00648-6},
   year = {2022},
   type = {Journal Article}
}

@inproceedings{RN105,
  title={Why should {I} trust you? Explaining the predictions of any classifier},
  author={Ribeiro, Marco Tulio and Singh, Sameer and Guestrin, Carlos},
  booktitle={Proceedings of the 22nd ACM SIGKDD International Conference on Knowledge Discovery and Data Mining},
  pages={1135--1144},
  year={2016}
}

\end{document}